\documentclass{article}
\usepackage{iclr2027_conference,times}

\usepackage{amsmath,amsfonts,bm}

\def\eqref#1{equation~\ref{#1}}

\def\1{\bm{1}}

\DeclareMathAlphabet{\mathsfit}{\encodingdefault}{\sfdefault}{m}{sl}
\SetMathAlphabet{\mathsfit}{bold}{\encodingdefault}{\sfdefault}{bx}{n}

\usepackage{hyperref}
\usepackage{url}
\usepackage{graphicx}
\usepackage{booktabs}
\usepackage{amsmath,amssymb}
\usepackage{float}
\usepackage{capt-of}
\usepackage{array}
\usepackage{tabularx}
\graphicspath{{figures/}}

\title{Useful to Whom?\\The Easy-versus-Coverage Boundary Moves with the Learner}

\iclrfinalcopy
\author{Yangze Liu \\
Shandong University \\
\texttt{yangze2@illinois.edu} \\
\And
Xiao-Long Yin \\
Nanjing University \\
\texttt{yinxl@lamda.nju.edu.cn} \\
\And
Zhongyi Han\thanks{Corresponding author.} \\
Shandong University \\
\texttt{zhongyi.han@sdu.edu.cn} \\
}

\begin{document}

\maketitle
\lhead{Preprint.}

\begin{abstract}
Is the same data equally useful to every learner? The question is whether the value of a training sample belongs to the sample or arises in its interaction with the learner. We test it with a selection experiment. Under a per-class budget, two kinds of samples compete for the slots: easy samples, which a short training run learns first, and coverage samples, which spread over the class in feature space. Easy samples win at small budgets and coverage samples at large ones, and the budget at which the winner flips, the crossover $m_0$, can be measured. We ask whether $m_0$ is set by the data or by the learner. To separate the two, we freeze the selected subsets and change only the learner that trains on them. On low-resolution ImageNet-100, doubling the width of a ResNet-18 moves the crossover from 57 to 85 samples per class on the same subsets. Enlarging the input grid from 32 to 64 pixels at fixed image information moves it from 85 to 57, and a single stride change reproduces or removes this shift. The grid shift shrinks as width grows. Under the native ImageNet-1k protocol, a $4\times$ range of width moves the boundary by at most a few samples per class. On a Vision Transformer, coverage wins at every measured budget, even with easy subsets that work on the convolutional learner. The budget at which one kind of sample stops being the more useful kind therefore belongs to the learner, and the preferred regime of a selection strategy has to be read against the learner that will use the data.
\end{abstract}

\section{Introduction}
\label{sec:intro}

Not all training data is equally useful, and a large literature selects the subset most worth training on \citep{welling2009herding,sener2018active,paul2021deep,guo2022deepcore}. In practice, usefulness is treated as a property of the sample. Scores are computed once, selected subsets are reused across models, and benchmarks report a winning strategy per dataset. Yet learners differ in what they can take from a sample. A convolutional network and a transformer impose different inductive biases and place the same image at different points of different feature spaces, and a wide network fits a small subset differently from a narrow one. Whether the same sample is worth the same to each of them is an open question, and the answer has consequences. If usefulness depends on the learner, a selection rule validated on one model carries no guarantee for the next, and a winner reported per dataset is a statement about one learner. This paper asks whether usefulness depends on who is doing the learning.

The question is hard to put to a test. Two learners differ in accuracy on any subset, so a difference in accuracy says nothing about whether the value of the data changed between them. What is needed is a quantity that describes the usefulness of the data and that a change of learner could, in principle, leave untouched. A second difficulty is a confound. Usefulness is usually scored by a short training run, and when the learner changes, the scorer changes with it, so a moving verdict could come from the scorer rather than from the learner.

Such a quantity is supplied by two kinds of samples that pull in opposite directions. Easy-first selection keeps the samples a short training run learns first. Coverage selection keeps the samples that spread over the class in feature space. Which kind wins depends on the budget: easy samples win when the budget is small, and coverage samples win when it is large \citep{sorscher2022beyond,liu2026dataside}. The budget at which the winner flips is the crossover $m_0$ of \citet{liu2026dataside}, who also found it stable on the data side: it stays roughly fixed when the sample pool shrinks. We freeze the selected subsets and change nothing but the learner that trains on them, which takes the scorer out of the comparison. If usefulness were a property of the data, $m_0$ would stay in place. If it depends on the learner, $m_0$ should move.

\begin{figure}[t]
\centering
\includegraphics[width=\textwidth]{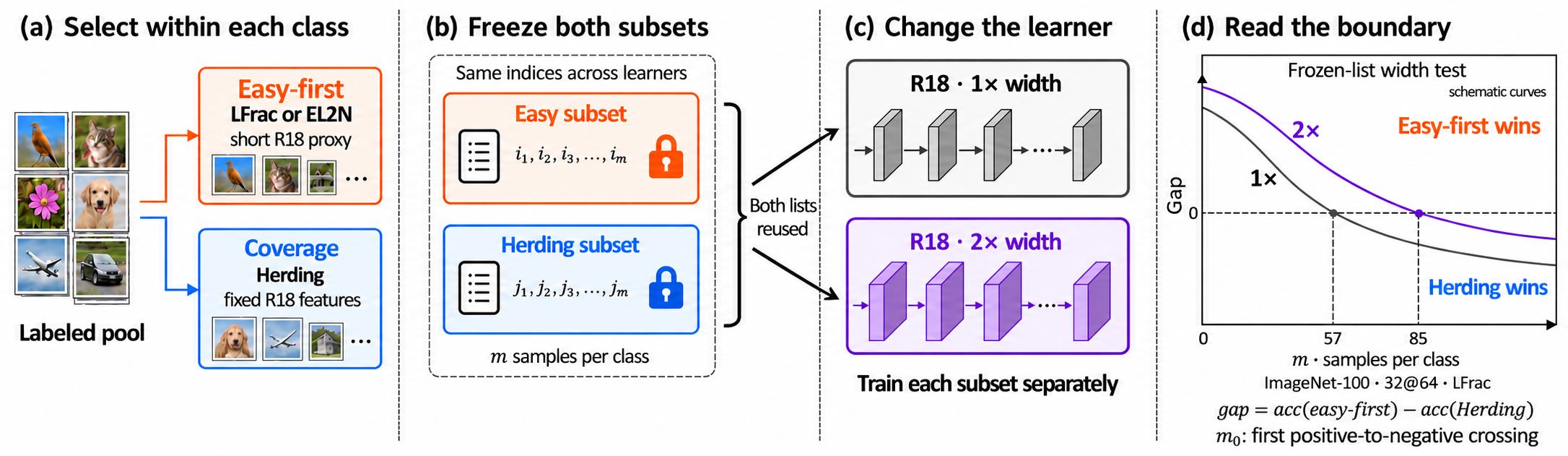}
\caption{\textbf{One instrument, one manipulation.} An easy-first set and a coverage set are selected within each class by ResNet-18 (\textbf{a}), both index lists are frozen (\textbf{b}), only the learner that trains on them changes (\textbf{c}), and the crossover $m_0$ is read from the gap between the two sets (\textbf{d}; schematic curves, measured crossings from Figure~\ref{fig:frozen}A). The grid, stride, and learner-family experiments change something else in \textbf{c}.}
\label{fig:design}
\end{figure}

\begin{figure}[t]
\centering
\includegraphics[width=\textwidth]{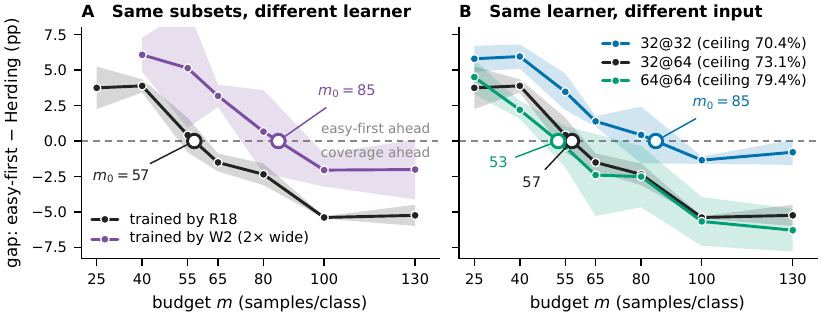}
\caption{\textbf{The crossover moves with the learner, not with the data.} Per-budget gap (easy-first $-$ Herding, percentage points) on low-resolution ImageNet-100, mean and $\pm1$ std over 3 seeds. \textbf{A}: the same frozen subsets, selected by ResNet-18 (R18) on the 32@64 cell, trained by R18 and by a $2\times$-wide variant (W2). \textbf{B}: R18 on three input cells; a@b means a px of retained information on a b-px input grid.}
\label{fig:frozen}
\end{figure}

With this one instrument, the frozen-subset crossover, we ask three questions in turn. Does the boundary move with the learner at all? Which part of the learner moves it? Does the movement survive at full scale? The headline answer is in Figure~\ref{fig:frozen}A: the same subsets, trained by a twice-wider ResNet-18, flip from easy-first to coverage at a budget half again as large. The paper makes three contributions.

\begin{itemize}
\item \textbf{Yes: which data is useful depends on the learner.} On frozen subsets, widening a ResNet-18 moves $m_0$ from 57 to 85 samples per class, so at the budgets in between the same two subsets favor opposite strategies on the two learners, and following the wrong learner's verdict costs about a tenth of the accuracy. On a Vision Transformer the easy set that serves the convolutional learner is worth no more than a random subset, and coverage wins at every budget tested.
\item \textbf{Inside a convolutional learner the lever is spatial, gated by capacity.} Enlarging the input grid at fixed image information moves $m_0$ far to the left. A single stride change with the grid fixed reproduces or removes the shift, so the carrier is the feature-map size inside the learner, and its last stage suffices. The shift shrinks as the learner widens.
\item \textbf{The boundary is not set by parameter count.} Under the native ImageNet-1k protocol, a $15\times$ range of parameters moves $m_0$ by a few samples per class, and richer input damps the width response on ImageNet-100 as well. The extrapolation that the crossover tracks samples per parameter fails. What sets it is how the learner processes space, not the scale of the learner.
\end{itemize}

\section{Related Work}
\label{sec:related}

Data selection supplies the instrument. Coverage methods pick representatives in feature space \citep{welling2009herding,sener2018active}. Training-dynamics methods score samples by how early or how reliably they are learned \citep{toneva2019empirical,swayamdipta2020dataset,paul2021deep,jiang2021characterizing}. Optimization-based methods match gradients or submodular objectives \citep{mirzasoleiman2020coresets,killamsetty2021gradmatch,iyer2021submodular}, proxy-model selection delegates the choice to a small model \citep{coleman2020selection}, and hybrid methods enforce coverage over difficulty strata \citep{zheng2023coverage,xia2023moderate}. \citet{liu2026benchmark} evaluates these families under one downstream protocol, and \citet{liu2026dataside} turns the easy-versus-coverage crossover into a measured quantity in absolute samples per class, introduces the LFrac probe, and fixes Herding as the coverage anchor. We adopt that framework unchanged. That work varies the data under a fixed learner. This paper varies the learner under fixed data and proposes no selection method.

Two lines of work already place the learner in the picture. Data valuation defines the worth of a sample relative to one fixed learning algorithm \citep{koh2017understanding,ghorbani2019data}. A second line reports that difficulty rankings computed on one architecture largely agree with those computed on another \citep{toneva2019empirical,hacohen2020agree,jiang2021characterizing,baldock2021difficulty}, and cross-architecture evaluations measure how much of a subset's advantage survives a change of network \citep{coleman2020selection,guo2022deepcore}. These lines ask whether scores and subsets port across learners. We ask a different question, whether the budget at which the winning kind of sample flips is itself set by the learner, and we answer it by manipulating the learner while the subsets stay fixed.

\citet{sorscher2022beyond} is the closest published work. They show that pruning by difficulty can beat power-law scaling and characterize the easy/hard flip line in the retained-fraction $\times$ data-quantity plane with perceptron theory, while keeping the learner fixed. By visual inspection their flip line lies near constant samples per parameter, which under naive extrapolation predicts that $m_0$ scales with width. Our frozen-subset width sweeps test that prediction directly.

\section{Experimental Setup}
\label{sec:setup}

\subsection{Measuring the crossover}
\label{sec:measure}

Easy-first subsets come from two learnability probes. LFrac scores each image by its mean correctness over the early epochs of a short-training ResNet-18 proxy, and keeps the highest scores. It follows \citet{liu2026dataside} and is the correctness statistic of Data Cartography \citep{swayamdipta2020dataset}. EL2N scores each image by the error norm $\lVert\mathrm{softmax}-\mathrm{onehot}\rVert_2$ at an early checkpoint, and keeps the lowest scores \citep{paul2021deep}. The two probes pick substantially different images: their easy sets overlap by about 17\%, below the 37--41\% self-overlap of LFrac rerun with a different seed (Appendix~\ref{app:e1}).

The coverage anchor is Herding, which greedily selects within each class so that the subset's feature mean approaches the class mean \citep{welling2009herding}. In a systematic evaluation under one downstream protocol, Herding is not first at every budget but stays reliably close to the best \citep{liu2026benchmark}, which makes it a stable comparison end across the whole budget range. Its features always come from the same 10-epoch ResNet-18, the fixed R18 extractor, whatever learner trains downstream.

For a budget of $m$ samples per class, write
\begin{equation}
\mathrm{gap}(m) = \mathrm{acc}(\text{easy-first}) - \mathrm{acc}(\text{Herding}).
\end{equation}
The crossover $m_0$ is the first positive-to-negative zero crossing of the gap, linearly interpolated between the two adjacent measured budgets, which form its measurement bracket. A gap that never changes sign within the measured range yields a censored reading, reported as a one-sided bound. A crossing whose adjacent gaps are within seed noise of zero is marked shallow. So defined, $m_0$ belongs to the whole measurement: the two selectors, the data, the learner, and the training protocol. Every experiment below holds the selectors, the data, and the protocol fixed and changes only the learner (Figure~\ref{fig:design}).

\subsection{Learner knobs and the frozen-subset design}
\label{sec:knobs}

We change the learner along three axes. \emph{Capacity}: ResNet-18 at $1\times$ width and its pure-width variants at $0.5\times$ (W05) and $2\times$ (W2), which halve or double the channel count of every layer, from 11.2M to 2.8M or 44.8M parameters, with depth, strides, and input grid unchanged (Figure~\ref{fig:design}c). \emph{Input grid}: we separate the information an image retains from the grid the learner receives, and write a@b for a px of retained information delivered on a b-px grid. Information is fixed by downsampling the native image to a px, and a larger grid is then produced by upsampling. The cells 32@32, 32@64, and 32@128 hold the information fixed and enlarge only the grid. The cells 32@64 and 64@64 hold the grid fixed and change only the information. With global average pooling, changing the grid changes no parameters, only feature-map sizes and compute, while changing width changes only channel and parameter counts. A \emph{stride} control separates the two halves of the grid change: it keeps the input grid and sets the stride of one layer to 1, or adds one at the stem, so that the feature maps change from that layer on while the image the learner receives does not (\S\ref{sec:grid}). \emph{Learner family}: a ViT-Tiny at patch size 8 or 4 on the 32@64 cell, trained with its own optimizer and schedule (Appendix~\ref{app:e2}).

Two designs differ in what the easy side sees. In the \emph{self-selected} design, each learner's easy set comes from a short-training proxy configured like the learner itself, while the coverage set always comes from the fixed R18 extractor. In the \emph{frozen-subset} design, the easy and coverage lists selected once by ResNet-18 are reused byte for byte, so the training learner is the only thing that changes. The frozen design carries every attribution claim in this paper. The self-selected design fills in the map around it.

\subsection{Data and protocol}
\label{sec:data}

Mechanism experiments run on ImageNet-100 under a low-resolution protocol with a $3\times3$-stem ResNet, 90 epochs of SGD, and a cosine schedule. Full-scale checks run on ImageNet-1k, on the same low-resolution ladder and under the native 224-px protocol with a $7\times7$-stem ResNet. Every load-bearing curve is a mean over 3 seeds, with proxy and downstream seeds paired so that error bars cover selection and training together. The ImageNet-1k low-resolution ladder is single-seed. Boundary differences between learners are given coarse 95\% intervals from an exact paired seed bootstrap (Appendix~\ref{app:uncertainty}). Training details, subset sizes, and the measurement status of every reported $m_0$ are collected in Appendix~\ref{app:appe}, and Appendix~\ref{app:inventory} lists every run.

\section{Results}
\label{sec:results}

\subsection{Does the boundary move with the learner?}
\label{sec:cross}

On frozen subsets, the boundary moves with the width of the learner. We take the easy set and the coverage set that ResNet-18 selected on the 32@64 cell, freeze both, and train W2 on them instead. Nothing else changes. ResNet-18 flips from easy-first to coverage at $m_0=57$ samples per class, and W2 flips on the same subsets later, at $m_0=85$ (Figure~\ref{fig:frozen}A). The paired seed bootstrap puts the shift at 28 samples per class and finds it positive in every draw (Appendix~\ref{app:uncertainty}). As a control, W2 with its own self-selected easy set flips at the same budget as on the frozen lists, so freezing the subsets is not what moved the boundary.

The shift separates two things that are easy to conflate. The subsets transfer: the frozen R18 lists serve W2 as well as W2's own lists do. What does not transfer is the verdict. Between the two crossings, the same two subsets favor coverage on ResNet-18 and easy-first on W2. Table~\ref{tab:cost} prices this. Following the narrower learner's verdict on the wider one loses about 3 points at 65 samples per class, a tenth of the accuracy there, and the loss has the same sign in every seed. Both strategies stay well above uniform sampling on both learners, so the choice is between two useful strategies, and the learner decides which one is more useful.

\begin{table}[t]
\centering
\small
\setlength{\tabcolsep}{5pt}
\caption{\textbf{What the wrong learner's verdict costs.} Test accuracy (\%, 3 seeds, std at most 2.5) of the same frozen R18-selected 32@64 subsets trained by R18 and by W2; Uniform is a seed-matched random subset. A learner's verdict at budget $m$ is the strategy with the higher accuracy on it; the cost rows give what one learner loses by following the other's verdict, computed before rounding.}
\label{tab:cost}
\begin{tabular}{llcccccc}
\toprule
Learner & Subset & $m$=40 & 55 & 65 & 80 & 100 & 130 \\
\midrule
R18 & easy-first & 27.1 & 31.6 & 33.7 & 36.1 & 37.6 & 42.7 \\
R18 & Herding & 23.3 & 31.2 & 35.2 & 38.5 & 43.0 & 48.0 \\
R18 & Uniform & 19.2 & 25.1 & 30.4 & 33.9 & 38.3 & 43.7 \\
W2 & easy-first (frozen) & 26.0 & 29.1 & 31.4 & 34.5 & 37.2 & 41.5 \\
W2 & Herding (frozen) & 19.9 & 23.9 & 28.2 & 33.9 & 39.3 & 43.6 \\
W2 & Uniform & 16.5 & 20.0 & 24.0 & 29.7 & 34.6 & 38.7 \\
\midrule
\multicolumn{2}{l}{Cost on W2 of R18's verdict} & 0 & 0 & 3.2 & 0.7 & 0 & 0 \\
\multicolumn{2}{l}{Cost on R18 of W2's verdict} & 0 & 0 & 1.5 & 2.4 & 0 & 0 \\
\bottomrule
\end{tabular}
\end{table}

The boundary moves right because the coverage set loses more of its value on the wider learner, not because the easy set gains any. Under the fixed schedule the wider learner is less accurate than ResNet-18 on every subset at every budget in Table~\ref{tab:cost}, although its full-data ceiling is higher (Appendix~\ref{app:b1}). The loss is unequal. The Herding arm gives up between 3.4 and 7.3 points, the easy-first arm never more than 2.5, and the uniform arm 2.7 to 6.4. The prototypical samples keep their worth across the width change and the spread-out samples do not. This is the specific sense in which the useful kind of data depends on the learner here.

\subsection{Which part of the learner moves it? The input grid}
\label{sec:grid}

The spatial size at which the learner processes an image moves the boundary far more than the information the image carries. The cells 32@32 and 32@64 hold the same 32-px information and differ only in the grid the learner receives. Enlarging the grid moves $m_0$ from 85 to 57 samples per class, and at 65 samples per class, between the two crossings, the two gaps have opposite signs beyond seed noise (Figure~\ref{fig:frozen}B). Raising the information instead, from 32@64 to 64@64, moves $m_0$ only to 53. The two knobs divide the labor. The grid moves the boundary and lifts the full-data ceiling by under 3 points. The information lifts the ceiling by over 6 points and leaves the boundary where it was (Appendix~\ref{app:b6}, Table~\ref{tab:w64}). TinyImageNet shows the same direction (Appendix~\ref{app:tiny}).

The grid effect passes the same frozen-subset test as width, and at the point estimate most of it belongs to the learner (Table~\ref{tab:gridcontrols}). We freeze the lists selected on 32@32 and train the 32@64 learner on them. The learner swap alone drops $m_0$ from 85 to 67, about two thirds of the full shift, and the reverse transplant gives the same accounting (Appendix~\ref{app:b3}). The remaining third travels with the lists, which do change with the grid: the easy lists of the two cells overlap by 40\%, against a random baseline of 6\%. Read as a $2\times2$ factorial, the learner main effect is twice the list main effect, with an interaction near zero at the point estimate, though the intervals are wide (Appendix~\ref{app:b3}). The lists move with the grid, but on this accounting the learner that reads them moves the boundary more.

\begin{figure}[t]
\centering
\includegraphics[width=\textwidth]{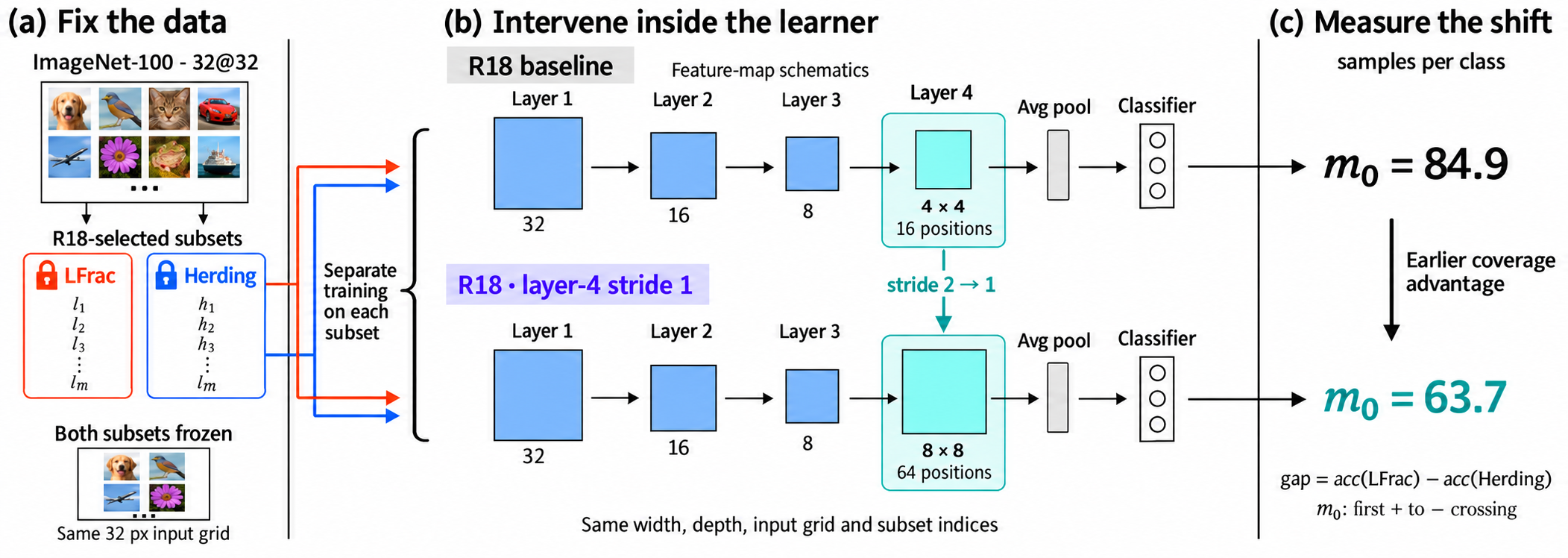}
\caption{\textbf{One stride change moves the boundary.} Frozen 32@32 lists trained by ResNet-18 (top) and by the same network with the stride of layer 4 set to 1 (bottom), which changes only the map entering global average pooling (Table~\ref{tab:gridcontrols}, layer-4 row).}
\label{fig:stride}
\end{figure}

A single stride change is enough to move the boundary (Table~\ref{tab:gridcontrols}, Figure~\ref{fig:stride}). Enlarging the grid doubles the side of every feature map. We keep the 32 grid and the frozen 32@32 lists and set the stride of one layer to 1, which doubles the maps from that layer on. Doing this at layer 2 moves $m_0$ from the native 85 down to 63, and doing it only at layer 4, which changes nothing but the map entering global average pooling, moves it to 64. The reverse control keeps the 64 grid and adds a stride at the stem, and $m_0$ returns to 90, past the native 32@32 reading. The carrier of the grid effect is therefore the size of the feature map inside the learner, and its last stage suffices.

The stride controls are also consistent with the transplants understating the learner's share. The reverse control lands at 90, past the native 32@32 reading, whereas the transplant of the same lists into the 32@32 learner stops at 76. Adding one stride removes the whole grid effect, not only the two thirds that the transplant attributed to the learner. Each control trains on the very images its lists were selected from, while a transplant trains the same lists at the other grid, so the comparison is consistent with part of the third that traveled with the lists being transplant friction rather than a property of the lists. The data do not separate the two.

The pooled map sets the boundary and the shape of the gap curve, but only part of the ceiling. The layer-4 change moves $m_0$ by the full learner-side share but lifts the full-data ceiling by about one point, less than half of the way to that of the 32@64 learner (Appendix~\ref{app:b6}), so the shift is not a by-product of a higher ceiling. Beyond the crossing the layer-2 control also leaves the 32@32 curve for the 64-grid one: at 100 samples per class and beyond its gaps lie between $-3$ and $-4.5$ points, past every native 32@32 reading and toward the 32@64 readings (Appendix~\ref{app:b1}). The lead that coverage builds at large budgets grows with the feature map as well.

\begin{table}[t]
\centering
\small
\caption{\textbf{Frozen-list controls on the grid axis.} Every row trains on the lists selected in the cell of the first column; only the learner changes. Transplants swap the learner's input grid; stride controls keep the grid and change one stride. The ladder lists the feature-map side after layers 1--4; $m_0$ in samples per class.}
\label{tab:gridcontrols}
\begin{tabular}{llcc}
\toprule
Lists from & Learner & Feature-map ladder & $m_0$ \\
\midrule
32@32 & 32@32 (native) & 32/16/8/4 & 84.9 \\
32@32 & 32@64 (transplant) & 64/32/16/8 & 66.6 \\
32@32 & 32@32, layer-2 stride 1 & 32/32/16/8 & 62.7 \\
32@32 & 32@32, layer-4 stride 1 & 32/16/8/8 & 63.7 \\
\midrule
32@64 & 32@64 (native) & 64/32/16/8 & 57.2 \\
32@64 & 32@32 (transplant) & 32/16/8/4 & 75.6 \\
32@64 & 32@64, stem stride 2 & 32/16/8/4 & 89.8 \\
\bottomrule
\end{tabular}
\end{table}

\subsection{Capacity gates the grid effect}
\label{sec:capgrid}

\begin{figure}[t]
\centering
\begin{minipage}[c]{0.50\textwidth}
\centering
\includegraphics[width=\linewidth]{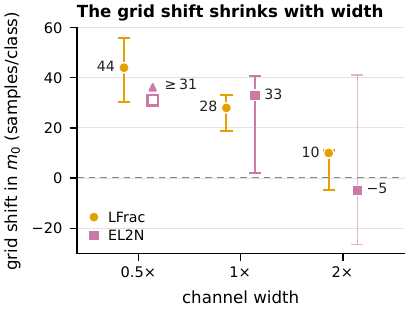}
\caption{\textbf{Capacity gates the grid effect.} The grid shift $m_0(32@32)-m_0(32@64)$ at each width for both probes, with coarse 95\% paired-seed intervals computed as in Appendix~\ref{app:uncertainty}; the hollow marker is a censored bound.}
\label{fig:capgrid}
\end{minipage}\hfill
\begin{minipage}[c]{0.46\textwidth}
\centering
\captionof{table}{\textbf{The capacity $\times$ grid map of $m_0$} on ImageNet-100, in samples per class. $^{\mathrm{s}}$: shallow crossing; $\ge$: censored.}
\label{tab:map}
\small
\begin{tabular}{lccc}
\toprule
 & $0.5\times$ & $1\times$ & $2\times$ \\
\midrule
\multicolumn{4}{l}{\emph{32@32}} \\
\quad LFrac & $\sim$108$^{\mathrm{s}}$ & 85 & $\sim$95$^{\mathrm{s}}$ \\
\quad EL2N & $\ge$100$^{\mathrm{s}}$ & 87 & 82 \\
\multicolumn{4}{l}{\emph{32@64}} \\
\quad LFrac & 64 & 57 & 85 \\
\quad EL2N & 69 & 54 & 87 \\
\bottomrule
\end{tabular}

\vspace{8pt}
\captionof{table}{\textbf{Width at 64@64,} the information-richest cell: self-selected LFrac lists against the fixed Herding set; $m_0$ with its measurement bracket, and full-data accuracy (\%, 3 seeds). Both outer crossings are shallow.}
\label{tab:w64}
\small
\begin{tabular}{lcccc}
\toprule
Width & Params & $m_0$ & Bracket & Full acc. \\
\midrule
$0.5\times$ & 2.8M & 57.1 & [55, 65] & 76.4 \\
$1\times$ & 11.2M & 53.0 & [40, 55] & 79.4 \\
$2\times$ & 44.8M & 59.1 & [55, 65] & 81.9 \\
\bottomrule
\end{tabular}
\end{minipage}
\end{figure}

Width and grid do not act independently. We cross three widths with the two grids and read $m_0$ in every cell with both probes (Table~\ref{tab:map}). The grid effect, the drop in $m_0$ from 32@32 to 32@64, is large at $0.5\times$ and $1\times$ width and shrinks at $2\times$, where the two grids read within one measurement bracket of each other (Figure~\ref{fig:capgrid}). The two probes agree on the whole map: the same cells are high, the same cells are low, and the shift collapses at $2\times$ for both, even though their easy sets share only about a sixth of their images. Along the 32@32 column the width ordering is not resolved, because both outer cells cross shallowly.

The interaction is not compute in disguise. Doubling the grid and doubling the width each raise the theoretical compute by about $4\times$. At $1\times$ width the grid doubling moves $m_0$ left by 28 samples per class, while at 32@64 the width doubling moves it right by the same 28. Equal compute increments move the boundary in opposite directions. Nor is it the ceiling: width lifts the full-data accuracy of the two grids by almost the same amount (Appendix~\ref{app:appb}), yet the interaction appears only at the boundary itself.

Nor does a single load scalar order the map. We pre-registered one, the input area divided by the width multiplier, and expected $m_0$ to fall as it grows. It does not. The cells $(0.5\times, 32@32)$ and $(2\times, 32@64)$ carry the same load and read about 108 and 85, and the highest-load cell, $(0.5\times, 32@64)$, reads 64, above the 57 of the $1\times$ cell (Appendix~\ref{app:b5}). We give no formula.

Richer input damps the width response in the same way that width damps the grid response. At 64@64, the information-richest low-resolution cell, the same three widths span a $16\times$ range of parameter count and lift the full-data ceiling by 5.5 points, yet $m_0$ stays between 53 and 59 samples per class, with overlapping brackets (Table~\ref{tab:w64}, Appendix~\ref{app:b4}). The width effect that dominates the 32@64 column is, at 64@64, within the resolution of the instrument.

\subsection{Changing the learner family removes the easy-first phase}
\label{sec:vit}

\begin{figure}[t]
\centering
\includegraphics[width=\textwidth]{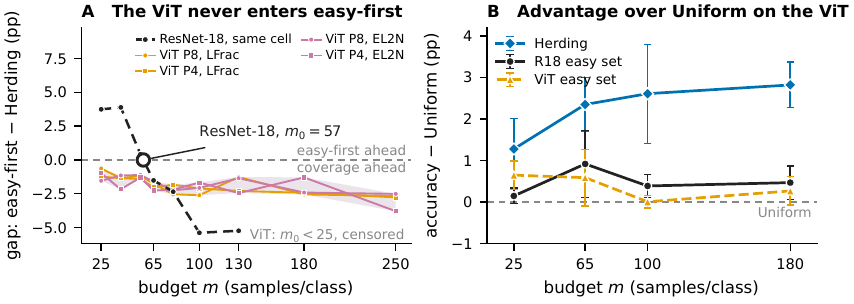}
\caption{\textbf{On a ViT, no easy-first phase within the measured range.} \textbf{A}: gap curves for ViT-Tiny at patch sizes 8 and 4 with both probes; the dashed curve is ResNet-18 on the same cell. \textbf{B}: accuracy advantage over uniform sampling on ViT-Tiny (patch 8) for the Herding set, the R18-selected easy set, and the ViT's own easy set; mean and std over 3 seeds.}
\label{fig:vit}
\end{figure}

On a ViT-Tiny, coverage wins at every measured budget, so the easy-first phase is missing from the measured range. Two patch sizes and two probes give four gap curves, and all four are negative from the smallest budget to the largest (Figure~\ref{fig:vit}A). By the convention of \S\ref{sec:measure} each reads as left-censored, $m_0<25$, whereas the LFrac crossings of convolutional learners at three widths on the same cell all fall between 57 and 85. A change of family removes a phase that no width setting did.

A transplant narrows the explanation. If the ViT merely scored its own easy set badly, an easy set of independent origin with proven small-budget value on the convolutional learner should restore the phase. It does not. The R18 easy set on the ViT gains only 0.1 to 0.9 points over uniform sampling, the same as the ViT's own easy set, while Herding leads uniform by 1 to 3 points across the range (Figure~\ref{fig:vit}B). The two lists overlap at the random baseline, so they share no images beyond chance (Appendix~\ref{app:c2}). Two easy sets of independent origin land at the uniform level on the same learner, so the missing easy-first phase belongs to the learner family and not to one scorer.

The missing phase is not a failure to fit. The ViT reaches 99.8\% training accuracy at 180 samples per class, yet its test accuracy at 250 stays near 30\%, far below the 49\% it reaches on the full data (Appendix~\ref{app:c2}). Patch size changes none of this: both patch sizes have nearly identical ceilings and the same four censored readings.

\subsection{Does it hold at full scale?}
\label{sec:in1k}

Under the native ImageNet-1k protocol, width moves the boundary far less than on ImageNet-100. We train the same frozen R18-selected subsets at three widths from $0.5\times$ to $2\times$ and read $m_0$ with both probes (Table~\ref{tab:in1k}, Appendix~\ref{app:d3}). Within each probe the three crossings fall in one sampled budget bracket, and the paired seed bootstrap separates the probes (Appendix~\ref{app:uncertainty}). Across the $4\times$ range of width, LFrac's boundary moves by under one sample per class, with an interval that includes zero, while EL2N's moves right by about seven, with an interval that excludes zero. Both are small next to the 28-sample shift of a single width doubling on ImageNet-100, while width lifts the accuracy of both strategies by several points per doubling (Appendix~\ref{app:d3}). At full scale, width changes how well the learner learns from either subset far more than which one to learn from.

\begin{figure}[t]
\centering
\includegraphics[width=\textwidth]{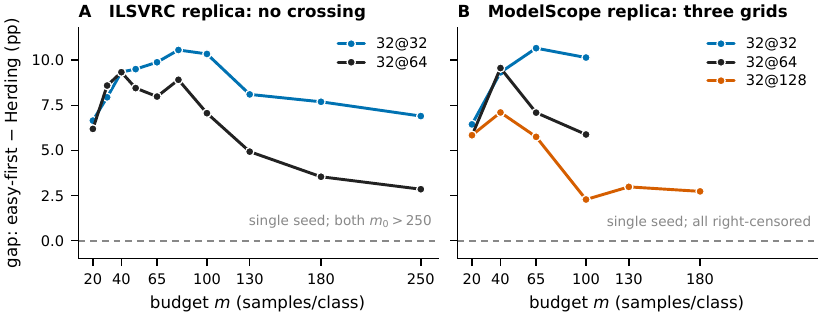}
\caption{\textbf{The ImageNet-1k low-resolution ladder yields orderings and censored bounds, not crossings.} Single seed. \textbf{A}: ILSVRC replica, two grids. \textbf{B}: ModelScope replica, three grids. The two replicas build their subsets independently and are compared only within a panel.}
\label{fig:scale}
\end{figure}

\begin{table}[H]
\centering
\small
\caption{\textbf{Native ImageNet-1k width sweep on frozen subsets.} $m_0$ per probe, in samples per class; within a probe the three crossings share one sampled bracket, [38.4, 51.2] for LFrac and [25.6, 38.4] for EL2N. Accuracy (\%, 3 seeds) at 64.1 samples per class.}
\label{tab:in1k}
\begin{tabular}{lccccc}
\toprule
 & & \multicolumn{2}{c}{$m_0$} & \multicolumn{2}{c}{Acc.\ at 64.1} \\
\cmidrule(lr){3-4}\cmidrule(lr){5-6}
Width & Params & LFrac & EL2N & Easy-first & Herding \\
\midrule
$0.5\times$ & 3.1M & 43.8 & 31.7 & 38.1 & 40.5 \\
$1\times$ & 11.7M & 42.5 & 34.0 & 42.3 & 45.0 \\
$2\times$ & 45.7M & 44.4 & 38.4 & 44.8 & 47.7 \\
\bottomrule
\end{tabular}
\end{table}

The grid axis cannot be measured at this scale, but it points the same way. On the low-resolution ladder on ImageNet-1k, easy-first stays ahead of Herding out to 250 samples per class at 32@32 and 32@64, and out to 180 at 32@128, so all three readings are censored (Figure~\ref{fig:scale}). The larger grid still erodes the lead faster. From its peak to 250 samples per class, the gap at 32@64 falls by about twice as much as at 32@32, and at 100 samples per class the gap decreases strictly with the grid across the three cells (Figure~\ref{fig:scale}B, Appendix~\ref{app:appd}). The ordering is the one ImageNet-100 predicts, and the crossover itself lies beyond the measured range.

The full-scale width sweep also answers the extrapolation from \citet{sorscher2022beyond}. Across a $15\times$ range of parameters, the boundary does not scale with parameter count. What damps the width response is richer input, as the 64@64 control of \S\ref{sec:capgrid} already showed on ImageNet-100.

\section{Discussion and Conclusion}
\label{sec:conclusion}

We asked whether the budget at which easy samples stop being the more useful kind belongs to the data or to the learner, and measured it with the selected subsets frozen. On the evidence here it belongs to the learner. Doubling the width of a ResNet-18 moves the crossover on the same subsets from 57 to 85 samples per class. Inside a convolutional learner the strongest lever is spatial: enlarging the feature maps moves the boundary left, one stride change reproduces or removes the shift, and widening the learner damps it. Replacing the learner family with a ViT removes the easy-first phase from the measured range altogether. At full scale, under the native ImageNet-1k protocol, a $15\times$ range of parameters moves the boundary by only a few samples per class, in a probe-dependent way. Together with the data-side stability reported by \citet{liu2026dataside}, these results place the dividing line between the two kinds of useful data on the learner side of the measurement.

We offer one reading of the stride controls as a hypothesis. The classification head averages the last feature map over its positions, and the layer-4 control changes only the number of positions averaged, from 16 to 64. If easy samples, which are prototypical, are also spatially redundant, extra positions add less new signal per easy sample than per coverage sample, and coverage pays off at a smaller budget. The hypothesis predicts that the size of the pooled map, not the input grid or the image information, orders $m_0$ across convolutional learners. The map so far agrees: every $1\times$ learner that pools 16 positions reads between 85 and 90, every $1\times$ learner that pools 64 reads between 53 and 67, and the two cells that share a pooled map but differ in image information, 32@64 and 64@64, read 57 and 53. The prediction has so far held.

Width, which adds channels rather than positions, lengthens the easy-first phase by a different route, and Table~\ref{tab:cost} shows how: the coverage set loses far more accuracy on the wider learner than the easy set does. One candidate reading is that doubling the channels enlarges the representation in which coverage has to be achieved, so a coverage set of fixed size spans less of the class and pays off later, while prototypical samples are learned by either learner. But the same doubling moves nothing at 64@64 or under the native protocol, so whatever width does, it does only when the input is information-poor, and we leave it open.

The scope of these conclusions is set by the instrument. The boundary is measured for two easy-first probes against one coverage anchor, Herding. Whether another coverage criterion gives the same map is untested. The headline cell shows the largest width response in our map: at 32@32 the crossings are shallow, and at 64@64 and under the native protocol the response is small. Low-resolution runs use a fixed 90-epoch schedule, so the boundary positions are those of this schedule. The 64@64 sweep and the ImageNet-1k ladder use self-selected lists, the ladder a single seed, so they map directions rather than isolate causes. The ViT result is a censored bound from a learner far below its ceiling, trained with a different optimizer, so it speaks to the learner family rather than to an isolated inductive-bias variable. Throughout, we measure the position of the boundary, not the accuracy of the subsets, and because the probes score with labels, the statement concerns selecting from a labeled pool rather than which images to annotate first.

The practical consequence is direct. A selection strategy's preferred budget regime must be read against the learner that will use the data. The same budget can fall left of one learner's boundary and right of another's, and the winning strategy flips accordingly. Within one family the list transfers and only the verdict changes. Across families the list itself loses its value. Usefulness is then not something a sample carries with it but a relation between the sample and a learner, and the question that follows is which properties of a learner predict what it needs. On the evidence here the size of the pooled feature map is one such property and parameter count is not. Different learners need different data, and what kind a given learner needs is a question about that learner.

\subsubsection*{Reproducibility statement}

All experiments use public datasets: ImageNet-100 (a fixed 100-class subset of ImageNet-1k), ImageNet-1k, and TinyImageNet. Selector implementations build on DeepCore \citep{guo2022deepcore}. The proxies, the fixed extractor, the BN-safe subset sizes, and the training configuration of every learner are in Appendix~\ref{app:appe}, and the measurement rules for $m_0$ are in Appendix~\ref{app:e4}. Every main-text number was re-aggregated from the stored per-run result files and checked against reference values; Appendix~\ref{app:inventory} lists the run inventory by experiment group. The paired-seed bootstrap of Appendix~\ref{app:uncertainty} is a standalone script over those files. Class lists, subset indices, run configurations, and analysis scripts will be released with the paper.

\subsubsection*{AI use statement}

In this work, we used generative AI tools to assist with writing and with figure drawing, as well as to implement standard components of the experimental pipeline and to assist with data analysis. Text was drafted and edited with AI assistance. We have reviewed all AI-assisted work: every reported number was recomputed by the authors from the stored run records, all text was read and approved by the authors, and every label and number in the figures was checked against the text. We take responsibility for the final content of this work, including text, claims or artifacts produced with the aid of generative AI.

\bibliography{refs}
\bibliographystyle{iclr2027_conference}

\appendix
\setcounter{figure}{0}\renewcommand{\thefigure}{A\arabic{figure}}\renewcommand{\theHfigure}{A\arabic{figure}}
\section{TinyImageNet Grid-Ladder Corroboration}
\label{app:tiny}

The TinyImageNet experiments examine whether the grid-direction effect replicates on another dataset; they are not load-bearing for the main text's causal identification. The data are natively 64px. Following the a@b information-on-grid notation in \S3.2, the ladder comprises 32@32 (downsampled), 64@64 (native), and 64@128 (upsampled from 64px), while the learner remains fixed. LFrac20 denotes the LFrac easy set obtained from a 20-epoch R18 short-training proxy. In the table, $f$ is the retained fraction per class, and each cell reports the test-accuracy gap between LFrac20 and fixed R18-Herding in percentage points.

\begin{table}[H]
\centering
\small
\resizebox{\textwidth}{!}{%
\begin{tabular}{lrrrrrrl}
\toprule
Grid & f=.05 & f=.10 & f=.20 & f=.30 & f=.50 & f=.70 & Crossover \\
\midrule
32@32 & +5.73 & +5.92 & +2.90 & +1.39 & +0.85 & +1.08 & No flip within range \\
64@64 & +3.48 & +2.14 & +0.85 & +0.39 & $-$0.51 & $-$0.10 & Approx.\ f=.39, i.e., approx.\ 193 samples/class \\
64@128 (upsampled) & +2.81 & $-$0.69 & $-$0.78 & $-$0.49 & --- & --- & Approx.\ f=.09, i.e., approx.\ 45 samples/class \\
\bottomrule
\end{tabular}%
}
\end{table}

The ordering is 64@128$<$64@64$<$32@32: the denser the grid received by the learner, the earlier geometric coverage takes the lead. The comparison between 64@64 and 64@128 fixes the native 64px information and changes only the grid received by the learner, providing pure grid corroboration in the same direction as the main results. By contrast, moving from 32@32 to 64@64 also restores information lost through downsampling and therefore cannot isolate the grid effect on its own.

These experiments use only one seed, and an independent standard-deviation check gives approximately 0.34 percentage points. For 64@128, only $f\in\{.05,.10,.20,.30\}$ is completed; the final two budgets are not measured. More importantly, the information side of 64@128 remains 64px and contains no new high-frequency information, so it cannot be interpreted as ``higher information resolution.'' We therefore do not use its absolute $m_0$ as load-bearing evidence in the main text and treat only its direction as cross-dataset corroboration.

\begin{figure}[H]
\centering
\includegraphics[width=0.7\textwidth]{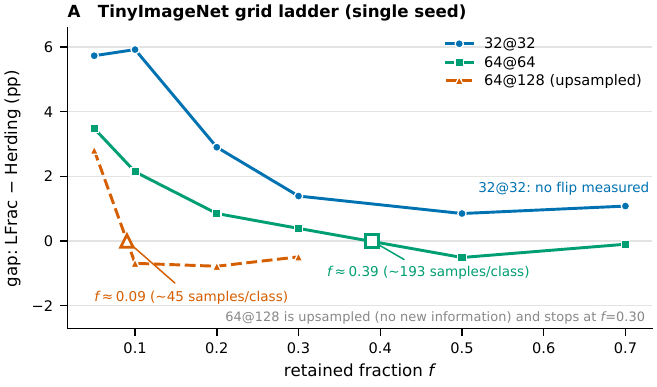}
\caption{\textbf{TinyImageNet grid ladder.} The x-axis shows the retained fraction $f$ per class, and the y-axis shows the LFrac20$-$Herding test-accuracy gap in percentage points. Open markers indicate linear zero crossings between adjacent measured points. The 64@128 condition is upsampled and stops after f=.30, while 32@32 does not flip within the maximum budget.}
\label{fig:tiny}
\end{figure}

\setcounter{figure}{0}\renewcommand{\thefigure}{B\arabic{figure}}\renewcommand{\theHfigure}{B\arabic{figure}}
\section{Complete State of the ImageNet-100 Capacity $\times$ Grid Study}
\label{app:appb}

\subsection{Six-Cell Map and Two Probes}
\label{app:b1}

All three widths belong to the same $3\times3$-stem \texttt{ResNet\_32x32} family. The parameter counts of $0.5\times$/$1\times$/$2\times$ are approximately 2.82M/11.2M/44.8M, respectively. Depth, stride, receptive field, and input grid remain fixed as width changes. All coverage sets are constructed using the fixed 10-epoch R18 extractor, and their feature source never changes with the training learner.

\begin{table}[H]
\centering
\scriptsize
\setlength{\tabcolsep}{3pt}
\begin{tabularx}{\textwidth}{l*{6}{>{\raggedright\arraybackslash}X}}
\toprule
Probe & $(0.5\times,32)$ & $(1\times,32)$ & $(2\times,32)$ & $(0.5\times,64)$ & $(1\times,64)$ & $(2\times,64)$ \\
\midrule
LFrac & Approx.\ 108; measured, shallow; [100,250) & 85; measured; [80,100] & Approx.\ 95; measured, shallow; [80,110] & 64; measured; [55,65] & 57; measured; [55,65] & 85; measured; [80,90] \\
EL2N & $\gtrsim100$; shallow crossing, lower bound only & 87; measured; the $m=100$ reading is coarse, conservative bracket [87,130) & 82; measured; [80,100] & 69; measured; [65,80] & 54; measured; [40,55] & 87; measured; [80,100], decisively confirmed at m130 \\
\bottomrule
\end{tabularx}
\end{table}

Except for EL2N at $(0.5\times,32)$, for which we report only a lower bound from a shallow crossing, every cell is measured. Cells marked ``shallow'' must be interpreted together with their stated measurement brackets. The LFrac and EL2N easy sets overlap by only approximately 16\%--17\%, yet yield the same capacity-gating verdict. For EL2N at $(0.5\times,32)$, the difference lower bound $\ge31$ follows from $m_0\ge100$ and 69 for grid 64 at the same width. The relative ordering of $0.5\times$ and $1\times$ cannot be resolved, and the coarse leftward shift supports only the direction and scale. Increasing capacity raises the full-data accuracy ceilings of 32@32 and 32@64 almost uniformly by 3.0 and 2.1 percentage points (\S\ref{sec:capgrid}, ceiling control).

The per-budget gap sequences for the two baseline cells at $1\times$ width are shown below (LFrac$-$Herding, mean$\pm$standard deviation over 3 seeds, in percentage points; the R18 curves in Figure~\ref{fig:frozen} and \S\ref{sec:grid} are plotted from these values). At $m=65$ the two gaps have opposite signs, each more than two standard deviations from zero. For the remaining cells, we report only the interpolated $m_0$ and measurement brackets in the table above. The densified readings around the W2 flips appear in Appendix~\ref{app:b2}, while the complete sequences for the transplant and the 64@64 width sweep appear in Appendices~\ref{app:b3} and~\ref{app:b4}.

\begin{table}[H]
\centering
\scriptsize
\resizebox{\textwidth}{!}{%
\begin{tabular}{lccccccccc}
\toprule
m & 25 & 40 & 55 & 65 & 80 & 100 & 130 & 180 & 250 \\
\midrule
$(1\times,32@32)$ & +5.80$\pm$0.92 & +5.96$\pm$0.85 & +3.47$\pm$1.29 & +1.39$\pm$0.37 & +0.43$\pm$2.00 & $-$1.35$\pm$0.26 & $-$0.79$\pm$0.89 & $-$2.39$\pm$0.53 & $-$2.92$\pm$0.82 \\
$(1\times,32@64)$ & +3.75$\pm$1.50 & +3.89$\pm$0.46 & +0.42$\pm$1.22 & $-$1.51$\pm$0.63 & $-$2.35$\pm$0.76 & $-$5.39$\pm$0.08 & $-$5.24$\pm$0.75 & $-$5.94$\pm$0.24 & $-$5.25$\pm$0.66 \\
\bottomrule
\end{tabular}}
\end{table}

\subsection{Densified W2 Budgets and Treatment of Shallow Crossings}
\label{app:b2}

The densified W2 budgets expose two distinct types of readings. For W2 in the 32@64 cell, the gap is $+1.97\pm0.10$ at m=80 and $-2.21\pm2.56$ at m=90. We therefore report $m_0$=85 with a flip bracket of [80,90]. In the cross arm, where W2 trains on the frozen R18 subsets (\S\ref{sec:cross}), the gap remains positive at m=80 and turns negative at m=100, yielding an interpolated $m_0$=85 and a flip bracket of [80,100]. For W2 in the 32@32 cell, the gaps at m=80/90/100/110/130 are +1.01/+2.81/$-$2.32/$-$0.71/$-$1.87, respectively. Because the readings near m=90 and m=100 are non-monotonic, we do not use the spurious precision produced by automatic interpolation. Instead, we report approximately 95, mark the crossing as shallow, and use the bracket [80,110], following the non-monotonicity widening rule in Appendix~\ref{app:e4}. For W2-EL2N, the $m_0$ values for input grids 32 and 64 are 82 and 87, each interpolated from a sign flip within the same pair of adjacent measured budgets, [80,100]; the difference of 5 samples/class is smaller than the width of this shared flip bracket. This describes budget resolution, not statistical equivalence.

The densified budgets use exact total sample counts to ensure that the final batch under batch 256 satisfies the BN constraint: m=90/110 use 9000/11000 samples, with remainders of 40/248, respectively. Under these settings, W2-Herding accuracies in the 32@64 cell at m=90/110 are $37.17\pm2.10$/$40.81\pm1.52$, and those in the 32@32 cell are $31.04\pm2.16$/$37.31\pm0.30$. The densified points calibrate and characterize the measurement brackets; they do not change the definition of Herding or the fixed R18 extractor.

\subsection{Cross-Grid Subset Transplant ($2\times2$)}
\label{app:b3}

The cross-grid transplant follows the same design as the cross-selector experiment in \S\ref{sec:cross}: the subsets in both arms are frozen, and only the input grid of the training learner changes. The easy set is the LFrac list from the 10-epoch proxy in the 32@32 cell, while the coverage set is the native Herding selection from the 32@32 cell, whose features still come from the fixed R18 extractor. The two cells share the same within-class sample pool and identical index ordering, so the lists can be transplanted byte for byte. The LFrac lists from the two cells overlap by 39.7\% (random baseline: 6.3\%), confirming that the lists do move with the grid and that the transplant is substantive.

\begin{table}[H]
\centering
\small
\begin{tabular}{lcc}
\toprule
List $\backslash$ Learner & 32@32 & 32@64 \\
\midrule
32@32 subset & 84.9 & 66.6 \\
32@64 subset & 75.6 & 57.2 \\
\bottomrule
\end{tabular}
\end{table}

The table reports interpolated $m_0$ values for the four combinations in samples/class. The diagonal contains the native readings for the two cells, while the off-diagonal contains the two transplants. The forward transplant (32@32 list$\to$32@64 learner) has a flip bracket of [65,80], with the following per-budget gaps (mean$\pm$standard deviation over 3 seeds): m25 $+5.48\pm0.93$ / m40 $+6.86\pm1.54$ / m55 $+3.48\pm4.25$ / m65 $+0.11\pm0.88$ / m80 $-0.88\pm2.58$ / m100 $-4.63\pm0.72$ / m130 $-4.39\pm0.31$ / m180 $-4.55\pm0.53$ / m250 $-4.79\pm0.56$. The reverse transplant (32@64 list$\to$32@32 learner) has the same flip bracket [65,80]: m25 $+3.91\pm1.54$ / m40 $+3.80\pm0.38$ / m55 $+1.57\pm1.12$ / m65 $+0.81\pm0.74$ / m80 $-0.34\pm0.53$ / m100 $-3.11\pm0.55$ / m130 $-1.92\pm1.30$ / m180 $-3.49\pm1.03$ / m250 $-3.83\pm0.53$.

The $2\times2$ readout gives learner main effects of 18.3/18.4 samples/class, list main effects of 9.3/9.4 samples/class, and an interaction of approximately 0.05, so the two effects are additive at the level of point estimates. The bracket edges are shallow in both transplants: forward m65 $+0.11\pm0.88$ and reverse m80 $-0.34\pm0.53$ are both indistinguishable from zero. The qualitative verdict is carried by points farther from zero: forward m100 $-4.63\pm0.72$ and reverse m40 $+3.80\pm0.38$. The residual of approximately 9 samples/class on the list side has a consistent magnitude in both directions and is attributable to the list moving with the grid; transplant friction is not excluded, and we make no directional assumption about it. We therefore phrase the attribution in the main text as the learner swap covering about two thirds of the grid effect at the point estimate, and do not claim that the effect sits purely on the learner side. An exact paired seed bootstrap over the four cells, with the conventions of Appendix~\ref{app:uncertainty}, gives a learner main effect of 18.3 with coarse 95\% interval [5.9, 20.4], a list main effect of 9.4 with [3.9, 22.8], and a learner share of the native difference of 0.66 with [0.32, 0.81]; both main effects are positive in all 27 draws. The interaction's interval, [$-19.2$, 6.2], spans zero, so additivity is a point-estimate observation only.

\subsection{The 64@64 Width Sweep}
\label{app:b4}

We extend the width sweep in \S\ref{sec:capgrid} to the information-richest 64@64 cell using the same protocol as in \S\ref{sec:capgrid}: the easy arm uses the LFrac list from each learner's own 10-epoch proxy, while the coverage arm uses the Herding coverage set built from the fixed R18 extractor. The $m_0$ values for $0.5\times$/$1\times$/$2\times$ are 57.1/53/59.1 samples/class, with flip brackets of [55,65]/[40,55]/[55,65], respectively; the brackets overlap. The full-data accuracy ceilings are $76.39\pm0.16$/$79.39\pm0.38$/$81.87\pm0.40$. Increasing width raises the ceiling monotonically by 5.5 percentage points, yet $m_0$ remains within a narrow band of approximately 6 samples/class. Both new cells have shallow crossings: at m=55, the gaps are $+0.26\pm1.39$ for $0.5\times$ and $+0.70\pm2.59$ for $2\times$. The relative ordering of the three point estimates is therefore not robust. These readings show a small spread relative to their shallow-crossing brackets; they establish neither statistical equivalence nor a U-shaped dependence on width.

The per-budget gap sequences are shown below (LFrac$-$Herding, mean$\pm$standard deviation over 3 seeds, in percentage points):

\begin{table}[H]
\centering
\scriptsize
\resizebox{\textwidth}{!}{%
\begin{tabular}{lccccccccc}
\toprule
m & 25 & 40 & 55 & 65 & 80 & 100 & 130 & 180 & 250 \\
\midrule
$(0.5\times,64@64)$ & +4.88$\pm$1.28 & +2.84$\pm$1.66 & +0.26$\pm$1.39 & $-$0.96$\pm$1.14 & $-$2.71$\pm$1.19 & $-$4.41$\pm$1.58 & $-$5.23$\pm$0.24 & $-$5.95$\pm$1.45 & $-$6.12$\pm$1.13 \\
$(2\times,64@64)$ & +5.88$\pm$1.44 & +3.64$\pm$0.18 & +0.70$\pm$2.59 & $-$1.01$\pm$1.11 & $-$2.89$\pm$1.42 & $-$7.43$\pm$2.28 & $-$4.52$\pm$2.32 & $-$7.40$\pm$1.47 & $-$7.03$\pm$1.78 \\
\bottomrule
\end{tabular}}
\end{table}

Remark: We pre-registered an exploratory hypothesis that the coverage-approximation error in the within-class latent space could predict the ordering of $m_0$ across widths. The confirmatory test does not support it, and we do not present this hypothesis as a result of the paper.

\subsection{Rejection of the Pre-Registered Load Law}
\label{app:b5}

We pre-registered a single-scalar law, $\mathrm{load}=(\text{input-grid side length})^2/\text{width multiplier}$, which predicts that $m_0$ changes monotonically with load. The 3-seed test rejects two predictions: $(0.5\times,32)$ and $(2\times,64)$, which have the same load, yield $m_0\approx108$ (shallow crossing) and 85, respectively; the highest-load condition, $(0.5\times,64)$, yields 64, which is not lower than 57 for $(1\times,64)$. Accordingly, the main text does not present a single formula (\S\ref{sec:capgrid}).

\subsection{Stride Controls: Recovering and Removing the Grid Effect}
\label{app:b6}

The transplants of Appendix~\ref{app:b3} localize the grid effect to the training learner without saying what in the learner carries it. Enlarging the input grid from 32 to 64 changes the learner in one structural way: every feature map doubles in side length, so layers 1--4 run at 64/32/16/8 instead of 32/16/8/4. Two stride controls test whether this feature-map ladder is the carrier, holding the input grid fixed and changing only one stride. The forward control keeps the 32 grid and sets the stride of layer 2 to 1, so that layers 2--4 run at 32/16/8, the 32@64 sizes from layer 2 on (layer 1 still runs at 32 rather than 64). The reverse control keeps the 64 grid and sets the stride of the stem convolution to 2, so that layers 1--4 run at 32/16/8/4, the 32@32 sizes from layer 1 on (the stem itself still reads the 64 grid). Both learners keep the parameter count of ResNet-18 (11.2M) and the training configuration of their cell, and each is trained on the frozen lists of its own cell, the same files as the transplants, so each control is the exact mirror of one transplant with the grid held fixed instead of swapped. A third control asks which rung of the ladder carries the effect: it keeps the 32 grid and sets only the stride of layer 4 to 1, so that layers 1--3 are identical to the 32@32 learner and only the map entering global average pooling doubles, from $4\times4$ to $8\times8$ (ladder 32/16/8/8).

\begin{table}[H]
\centering
\scriptsize
\setlength{\tabcolsep}{3pt}
\begin{tabular}{llllccc}
\toprule
Control & Grid & Ladder (layers 1--4) & Lists & $m_0$ [bracket] & Native cell & Transplant, same lists \\
\midrule
Forward: layer-2 stride 1 & 32 & 32/32/16/8 & 32@32 & 62.7 [55,65] & 84.9 (32@32) & 66.6 (32@64 learner) \\
Reverse: stem stride 2 & 64 & 32/16/8/4 & 32@64 & 89.8 [80,100] & 57.2 (32@64) & 75.6 (32@32 learner) \\
Rung: layer-4 stride 1 & 32 & 32/16/8/8 & 32@32 & 63.7 [55,65] & 84.9 (32@32) & 66.6 (32@64 learner) \\
\bottomrule
\end{tabular}
\end{table}

Per-budget gaps (LFrac$-$Herding, mean$\pm$standard deviation over 3 seeds) are, for the forward control, m25 $+4.98\pm2.03$ / m40 $+4.69\pm2.92$ / m55 $+1.28\pm2.31$ / m65 $-0.38\pm0.92$ / m80 $-0.94\pm1.13$ / m100 $-4.25\pm1.60$ / m130 $-2.95\pm0.69$ / m180 $-4.38\pm1.01$ / m250 $-4.47\pm0.49$, and for the reverse control, m25 $+4.25\pm0.74$ / m40 $+4.57\pm0.74$ / m55 $+4.12\pm1.09$ / m65 $+3.17\pm1.89$ / m80 $+1.57\pm2.13$ / m100 $-1.63\pm0.44$ / m130 $-2.53\pm1.18$ / m180 $-2.69\pm0.29$ / m250 $-3.53\pm1.21$, and for the layer-4 control, m25 $+4.85\pm1.16$ / m40 $+5.71\pm1.40$ / m55 $+1.99\pm1.87$ / m65 $-0.29\pm1.44$ / m80 $-0.97\pm1.84$ / m100 $-3.38\pm0.55$ / m130 $-2.23\pm0.12$ / m180 $-3.45\pm1.32$ / m250 $-4.15\pm0.04$.

Both controls move the boundary to the side the ladder predicts. The forward control reads 62.7 against the 32@64 learner's 66.6 on the same lists, lower in each of the three seeds (80/63/51 against 84/77/58), so removing one stride reproduces what the 64-grid learner does to these lists; the whole gap curve takes the 64-grid shape, with the gaps at $m\ge100$ at $-4.2$ to $-4.5$ where the native 32@32 cell reads $-1.4$ to $-2.9$. The reverse control reads 89.8 against the 32@32 learner's 75.6 on the same lists, higher in each seed (93/87/81 against 77/65/74), and at $m\ge100$ its accuracies sit within a point of the native 32@32 cell's. Adding one stride at the 64 grid therefore removes the whole grid effect, not only the learner-side share the transplant isolated: the reverse control lands at the native 32@32 reading (84.9) rather than at the transplant's 75.6. The stem's finer view of the 64 grid, the only thing separating the reverse control from the 32@32 learner, does not hold the boundary at 57. The grid effect travels with the feature-map ladder. The layer-4 control then locates the rung: it reads 63.7 [55,65] (seeds 74/63/55 against the native 71/94/83), within a point of the layer-2 control's 62.7 and with the same gap shape at $m\ge100$, while its accuracy level stays within a point of the native cell's at every budget; in the paired bootstrap it sits below the native reading with $P=0.96$ and is indistinguishable from the layer-2 control. Doubling only the map that enters the pooled head, from 16 positions to 64, moves the boundary by the whole learner-side share; enlarging the earlier rungs as well adds no measurable shift beyond it. The full-data ceilings move the same way: the forward control reaches 73.0 (three seeds 73.1/73.0/73.0) where its own cell's learner reaches 70.4 and the 32@64 learner 73.1, and the reverse control reaches 69.1 (69.5/68.4/69.5) where its own cell's learner reaches 73.1 and the 32@32 learner 70.4. The layer-4 control reaches 71.5 (71.8/71.3/71.4), between its own cell's 70.4 and the 32@64 learner's 73.1: the last rung carries the whole boundary shift but only part of the small ceiling gain. Both controls also land nearer the other cell's native reading than the transplant with the same lists does (62.7 against 66.6, toward 57.2; 89.8 against 75.6, toward 84.9), which is consistent with part of the transplant's list-side share being friction: each control trains on the very images its list was selected from, whereas a transplant trains on the same images at the other grid. Both matches are structural rather than exact, since the forward control's layer 1 and the reverse control's stem still see the grid at their own cell's scale. The layer-4 control shows that the last rung is sufficient; whether an earlier rung would also suffice on its own, with the pooled map held at $4\times4$, is not tested.

\setcounter{figure}{0}\renewcommand{\thefigure}{C\arabic{figure}}\renewcommand{\theHfigure}{C\arabic{figure}}
\section{Four ViT Results and Transplant Control}
\label{app:appc}

\subsection{Four Left-Censored Measurements}
\label{app:c1}

ViT-Tiny P8 and P4 both use the 32@64 input grid and share the same data with the convolutional learners. They form $8\times8=64$ and $16\times16=256$ tokens, respectively, and both have an embed dimension of 192. The two patch sizes combined with LFrac/EL2N yield four measurement configurations. All four gaps are already negative at m=25, and easy-first never takes the lead at any measured budget.

\begin{table}[H]
\centering
\small
\begin{tabular}{llrrl}
\toprule
Learner & Probe & m=25 gap & m=250 gap & $m_0$ status \\
\midrule
ViT-Tiny P8 & LFrac & $-$0.63 & $-$2.66 & $<$25, left-censored \\
ViT-Tiny P8 & EL2N & $-$1.54 & $-$2.51 & $<$25, left-censored \\
ViT-Tiny P4 & LFrac & $-$1.23 & $-$2.77 & $<$25, left-censored \\
ViT-Tiny P4 & EL2N & $-$0.98 & $-$3.79 & $<$25, left-censored \\
\bottomrule
\end{tabular}
\end{table}

For LFrac, we show all nine measured budgets. For EL2N, we report only the endpoints and the censored status that the gap remains negative throughout.

\begin{table}[H]
\centering
\scriptsize
\resizebox{\textwidth}{!}{%
\begin{tabular}{lccccccccc}
\toprule
m & 25 & 40 & 55 & 65 & 80 & 100 & 130 & 180 & 250 \\
\midrule
P8 LFrac$-$Herding & $-$0.63 & $-$1.35 & $-$1.17 & $-$1.75 & $-$2.54 & $-$2.61 & $-$1.26 & $-$2.55 & $-$2.66 \\
P4 LFrac$-$Herding & $-$1.23 & $-$1.35 & $-$1.41 & $-$2.03 & $-$1.83 & $-$2.11 & $-$2.29 & $-$2.43 & $-$2.77 \\
\bottomrule
\end{tabular}}
\end{table}

All four easy-first methods remain close to Uniform. The LFrac ranges for P8/P4 are $-$0.03--+0.86 and $-$0.29--+0.21, respectively, while the EL2N ranges for P8/P4 are $-$0.26--+0.84 and $-$0.81--+1.01. The full-data accuracy ceilings of P8/P4 are 49.14\%/49.39\%, which are nearly identical. The censoring therefore cannot resolve whether patch granularity moves the exact $m_0$.

\begin{figure}[H]
\centering
\includegraphics[width=0.7\textwidth]{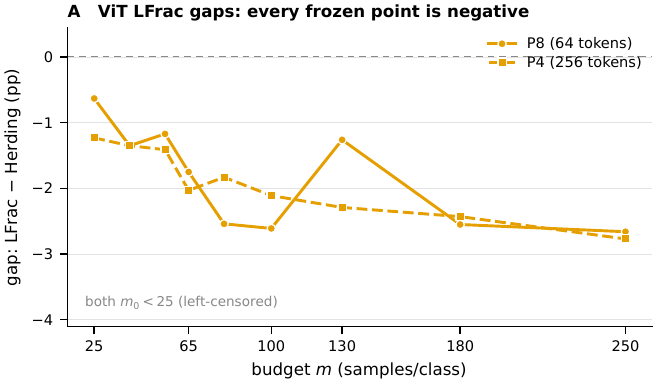}
\caption{\textbf{Complete LFrac gap sequences for ViT.} All nine measured points for both P8 and P4 lie below zero, so both can only be reported as $m_0<25$. Local rebounds in the curves do not change the censoring verdict.}
\label{fig:vitlfrac}
\end{figure}

\subsection{Transplanting the R18 Easy Set to ViT}
\label{app:c2}

The transplant control fixes the ViT-P8 learner and compares the R18-proxy easy set, the ViT self-selected easy set, Uniform, and fixed R18-Herding. The table reports the mean$\pm$standard deviation of test accuracy over 3 seeds.

\begin{table}[H]
\centering
\scriptsize
\begin{tabular}{lccccc}
\toprule
m & R18 easy set$\to$ViT & Uniform$\to$ViT & ViT easy set$\to$ViT & Herding$\to$ViT & R18 easy$-$Uniform \\
\midrule
25 & 10.4$\pm$0.1 & 10.2$\pm$0.1 & 10.9$\pm$0.3 & 11.5$\pm$0.8 & +0.1 \\
65 & 15.4$\pm$0.3 & 14.5$\pm$0.5 & 15.1$\pm$0.3 & 16.9$\pm$0.4 & +0.9 \\
100 & 18.4$\pm$0.7 & 18.0$\pm$0.4 & 18.0$\pm$0.3 & 20.6$\pm$1.0 & +0.4 \\
180 & 23.8$\pm$0.3 & 23.3$\pm$0.2 & 23.6$\pm$0.5 & 26.2$\pm$0.7 & +0.5 \\
\bottomrule
\end{tabular}
\end{table}

At the remaining measured budgets m=40/55/80, the gaps between the R18 easy set and Uniform are +0.51/+0.25/+0.33. Both the R18 and ViT easy-set curves remain close to Uniform, while Herding consistently leads by approximately 1--3 percentage points. The ViT and R18 easy sets overlap by approximately 5.6\%, close to the random baseline of 5.1\%. If the failure arose only because ViT's own scoring function failed, the R18 easy set validated on convolutional learners should restore the gain. It does not, ruling out this explanation. However, the ViT results still cannot be treated as precise boundary measurements: accuracy at m=250 is only approximately 28\%--30\%, far below the ceiling of approximately 49\%. The training subset is nonetheless fully fit (training accuracy 99.8\% at m=180), so the shortfall is not a failure to fit the training data. Whether a different recipe or pretraining would lift the ViT's small-budget accuracy, and with it the censoring verdict, is not tested.

\setcounter{figure}{0}\renewcommand{\thefigure}{D\arabic{figure}}\renewcommand{\theHfigure}{D\arabic{figure}}\setcounter{table}{0}\renewcommand{\thetable}{D\arabic{table}}\renewcommand{\theHtable}{D\arabic{table}}
\section{Full-Scale ImageNet-1k Results}
\label{app:appd}

\subsection{Low-Resolution Grid Ladder and Two Data Replicas}
\label{app:d1}

The low-resolution ladder fixes 32px information and compares 32@32, 32@64, and 32@128, all using the $3\times3$-stem \texttt{ResNet\_32x32}. We use two independent data replicas, ILSVRC and ModelScope. Their within-class index orders differ, so subsets cannot be reused across replicas; each replica constructs its subsets independently.

The gap curves of both replicas are plotted in Figure~\ref{fig:scale} of the main text.

Single-seed sweep on the ILSVRC replica:

\begin{table}[H]
\centering
\scriptsize
\resizebox{\textwidth}{!}{%
\begin{tabular}{lcccccccccc}
\toprule
m & 20 & 30 & 40 & 50 & 65 & 80 & 100 & 130 & 180 & 250 \\
\midrule
32@32 LFrac$-$Herding & +6.65 & +7.94 & +9.34 & +9.50 & +9.88 & +10.56 & +10.34 & +8.10 & +7.69 & +6.90 \\
32@64 LFrac$-$Herding & +6.19 & +8.59 & +9.33 & +8.45 & +7.98 & +8.91 & +7.06 & +4.93 & +3.54 & +2.85 \\
\bottomrule
\end{tabular}}
\end{table}

Single-seed sweep on the ModelScope replica:

\begin{table}[H]
\centering
\small
\begin{tabular}{lccccc}
\toprule
m & 20 & 40 & 65 & 80 & 100 \\
\midrule
32@32 LFrac$-$Herding & +6.44 & +9.32 & +10.66 & +9.99 & +10.14 \\
32@64 LFrac$-$Herding & +5.82 & +9.56 & +7.09 & +7.43 & +5.88 \\
\bottomrule
\end{tabular}
\end{table}

On the second seed of the ModelScope replica, the 32@32/32@64 gaps at m=100 are +9.48/+7.79. Across the four replicated points at m=40/100 in the two cells, every absolute cross-seed difference is below 2 percentage points. This supports the direction that ``32@64 erodes faster,'' but is insufficient to elevate the single-seed curves into precise $m_0$ measurements. We therefore report only $m_0>250$ for both 32@32 and 32@64.

\subsection{Shared Budgets and Tail for 32@128}
\label{app:d2}

The 32@128 gap rebounds from +2.28 at m=100 to +2.98/+2.73 at m=130/180, so it does not form an interpolable flip. We therefore report only $m_0>180$.

\begin{table}[H]
\centering
\caption{\textbf{Shared budgets across the three ImageNet-1k input grids.} Single-seed results; m is the absolute number of samples per class, and gaps are in percentage points.}
\label{tab:d1}
\small
\begin{tabular}{lccc}
\toprule
m & 32@32 gap & 32@64 gap & 32@128 gap \\
\midrule
20 & +6.44 & +5.82 & +5.84 \\
40 & +9.32 & +9.56 & +7.10 \\
65 & +10.66 & +7.09 & +5.75 \\
100 & +10.14 & +5.88 & +2.28 \\
\bottomrule
\end{tabular}
\end{table}

\begin{table}[H]
\centering
\caption{\textbf{Tail readings for ImageNet-1k 32@128.} Single-seed results; m is the absolute number of samples per class, test accuracy is in \%, and gaps are in percentage points.}
\label{tab:d2}
\small
\begin{tabular}{lcccc}
\toprule
32@128, m & LFrac acc. & Herding acc. & Uniform acc. & gap \\
\midrule
100 & 25.99 & 23.71 & 18.95 & +2.28 \\
130 & 29.12 & 26.14 & 22.21 & +2.98 \\
180 & 32.03 & 29.30 & 26.99 & +2.73 \\
\bottomrule
\end{tabular}
\end{table}

Across the four shared budgets, the gap generally approaches zero as the grid becomes denser. Together with the tail rebound, this ladder supports only an ordering of erosion rates and censored lower bounds.

\subsection{Capacity Sweep Under the Native-224px Protocol}
\label{app:d3}

All native experiments use the standard $7\times7$-convolution stem of \texttt{ResNet\_224wide}. Base widths 32/64/128 correspond to $0.5\times$/$1\times$/$2\times$, with parameter counts of 3.06M/11.69M/45.69M. Within each learnability probe, the three widths train on the same frozen R18 easy sets and share the fixed R18-Herding coverage set, making width the only change within the probe. Across the six shared points between W1 and the existing reference runs, the maximum absolute difference is 0.24 percentage points: +0.05/$-$0.02/+0.02 for LFrac and +0.24/$-$0.20/+0.03 for Herding. These differences are an order of magnitude smaller than the gaps measured in this section, which are approximately 0.5--5.8 percentage points. We therefore treat the readings from the two harnesses as following the same protocol.

LFrac20$-$Herding gap, mean over 3 seeds:

\begin{table}[H]
\centering
\scriptsize
\resizebox{\textwidth}{!}{%
\begin{tabular}{lcccccccc}
\toprule
samples/class & 12.8 & 25.6 & 38.4 & 51.2 & 64.1 & 76.9 & 89.7 & 102.5 \\
\midrule
W05($0.5\times$) & +5.80 & +3.19 & +0.91 & $-$1.28 & $-$2.35 & --- & --- & --- \\
W1($1\times$) & --- & --- & +0.48 & $-$1.04 & $-$2.71 & --- & --- & --- \\
W2($2\times$) & --- & +4.22 & +1.08 & $-$1.23 & $-$2.82 & $-$3.74 & $-$4.49 & $-$4.91 \\
\bottomrule
\end{tabular}}
\end{table}

The corresponding $m_0$ values are 43.8/42.5/44.4, and all three share the measurement bracket [38.4,51.2].
\begin{figure}[H]
\centering
\includegraphics[width=\textwidth]{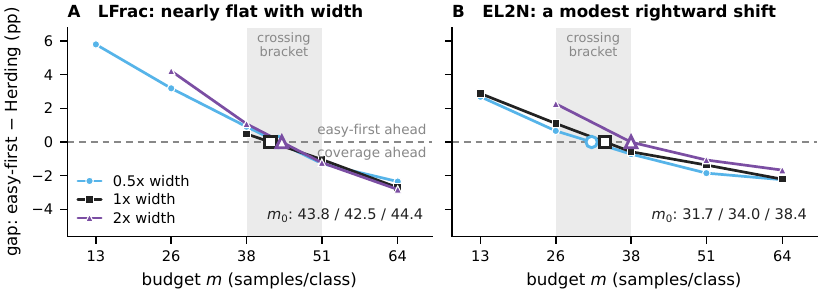}
\caption{\textbf{Native ImageNet-1k: per-budget gap at three widths on the same frozen subsets}, for LFrac (\textbf{A}) and EL2N (\textbf{B}). The gray band is the sampled budget bracket shared by the three crossings, not a confidence interval.}
\label{fig:in1kcap}
\end{figure}
 The EL2N20$-$Herding gaps are:

\begin{table}[H]
\centering
\small
\begin{tabular}{lccccc}
\toprule
samples/class & 12.8 & 25.6 & 38.4 & 51.2 & 64.1 \\
\midrule
W05($0.5\times$) & +2.69 & +0.66 & $-$0.72 & $-$1.85 & $-$2.24 \\
W1($1\times$) & +2.88 & +1.10 & $-$0.58 & $-$1.37 & $-$2.21 \\
W2($2\times$) & --- & +2.27 & $-$0.01 & $-$1.07 & $-$1.67 \\
\bottomrule
\end{tabular}
\end{table}

The corresponding $m_0$ values are 31.7/34.0/38.4, and all three share the measurement bracket [25.6,38.4]. W2 is $-0.01\pm0.10$ at 38.4, placing its flip at the edge of the bracket, so we report $m_0\approx$38--40. The two learnability probes have a systematic offset in absolute $m_0$, and different width responses: the EL2N rightward shift persists in the paired seed analysis below, whereas the LFrac endpoint difference is small and its interval includes zero.

Width moves the accuracy level. At 64.1 samples/class, LFrac-arm accuracies for $0.5\times$/$1\times$/$2\times$ are 38.14/42.30/44.83, while Herding-arm accuracies are 40.49/45.01/47.65. Each increase in width raises both arms by 2.5--4.5 percentage points in near synchrony, with smaller changes in their gap (\S\ref{sec:in1k}).

\subsection{Paired Seed Uncertainty for Frozen-Subset Width Contrasts}
\label{app:uncertainty}

We quantify uncertainty in boundary \emph{differences} using the existing three physical seeds. For each contrast, we enumerate all $3^3=27$ ordered draws of three seeds with replacement. A draw is shared across the two learners, both selectors, and every measured budget, preserving their pairing. For each learner we average the drawn gap curves and interpolate the first positive-to-negative crossing on its published budget grid, then subtract the two crossings. The table gives equal-tailed 95\% empirical percentile intervals using the inverse empirical CDF. There are only ten distinct seed multisets, so these are coarse exploratory uncertainty estimates, not equivalence tests or additional independent experiments. Draws without an observed crossing are counted separately; intervals for those rows are conditional on both crossings being observed. The reproducible analysis is supplied as \texttt{mech\_m0\_uncertainty.py}.

\begin{table}[H]
\centering
\small
\caption{\textbf{Width-induced shifts in $m_0$, in samples/class.} Positive means a rightward shift. ``Unresolved'' counts bootstrap draws without two observed crossings, out of 27.}
\begin{tabular}{llrrr}
\toprule
Setting / probe & Width contrast & Shift & 95\% interval & Unresolved \\
\midrule
IN-100 32@64 / LFrac & $1\times\to2\times$ & 27.7 & [22.4, 35.8] & 0 \\
IN-1k native / LFrac & $0.5\times\to1\times$ & $-1.2$ & [$-3.0$, $-0.8$] & 1 \\
 & $1\times\to2\times$ & 1.9 & [$-0.3$, 4.0] & 1 \\
 & $0.5\times\to2\times$ & 0.7 & [$-1.4$, 2.5] & 0 \\
IN-1k native / EL2N & $0.5\times\to1\times$ & 2.3 & [1.4, 2.4] & 0 \\
 & $1\times\to2\times$ & 4.4 & [3.8, 6.6] & 0 \\
 & $0.5\times\to2\times$ & 6.7 & [5.5, 8.7] & 0 \\
\bottomrule
\end{tabular}
\end{table}

The IN-100 frozen-subset shift remains positive in every bootstrap draw. Native ImageNet-1k gives a different result for each probe: the LFrac endpoint interval includes zero, while the EL2N shift is positive in every draw and in each physical seed. The missing crossing in one LFrac W1 draw occurs because that mean curve starts negative at its smallest measured budget; it is left-censored rather than extrapolated. These results support a smaller native-protocol width response, with a nonzero EL2N shift, and supersede a zero-effect interpretation based solely on shared budget brackets.

\setcounter{table}{0}\renewcommand{\thetable}{E\arabic{table}}\renewcommand{\theHtable}{E\arabic{table}}
\section{Experimental and Interpretation Protocols}
\label{app:appe}

\subsection{Selectors and Short-Training Proxies}
\label{app:e1}

LFrac takes the mean per-image correctness over the early epochs of the short-training proxy and selects the highest-scoring samples within each class. EL2N uses $\lVert\mathrm{softmax}-\mathrm{onehot}\rVert_2$ and selects the lowest-scoring samples within each class. LFrac and EL2N on ImageNet-100 use 10-epoch proxies. The ImageNet-1k low-resolution ladder uses LFrac10. Both LFrac and EL2N in the native-224px capacity sweep use the same fixed 20-epoch R18 proxy. The TinyImageNet corroboration uses LFrac20. The selector implementations build on DeepCore \citep{guo2022deepcore}.

In the ImageNet-100 capacity $\times$ grid experiments, the easy sets selected by LFrac and EL2N overlap by only approximately 17\% (random baseline: approximately 5\%), below the self-overlap of approximately 37\%--41\% obtained by rerunning the same LFrac configuration with only the seed changed. The difference between the lists therefore exceeds the seed noise of a single method, so the two learnability probes are substantially different measurements rather than one list twice (\S3.1).

Herding greedily matches the feature mean within each class. In every experiment in this paper, Herding features come from the fixed 10-epoch R18 extractor, even when the training learner is W05, W2, or ViT. The cross-selector experiment additionally freezes the R18 easy and coverage sets, changing only the downstream training learner. Uniform always samples uniformly within each class.

\subsection{Learners and Training Configurations}
\label{app:e2}

All low-resolution ResNet experiments use the $3\times3$-stem \texttt{ResNet\_32x32}, trained with SGD (momentum 0.9, Nesterov, weight decay 5e$-$4) and \texttt{RandomCrop(grid,pad=grid//8)+Flip} augmentation. The low-resolution downstream runs on ImageNet-100 and ImageNet-1k use batch 256 and 90 training epochs, with a cosine learning rate decreasing from 0.1 to 1e$-$4; the TinyImageNet corroboration uses batch 128 and 200 training epochs, everything else unchanged. The optimizer and training schedule remain fixed within each controlled comparison. The three stride controls of Appendix~\ref{app:b6} are the same \texttt{ResNet\_32x32} with one stride changed (layer 2 or layer 4 set to stride 1 at the 32 grid; the stem convolution set to stride 2 at the 64 grid), trained with the configuration of their cell. The native-224px capacity sweep uses the $7\times7$-stem \texttt{ResNet\_224wide}, batch 256, and 90 training epochs, with a cosine learning rate decreasing from 0.1 to 1e$-$4. The low-resolution and native-224px settings belong to two distinct stem families, and every load-bearing comparison remains within a family.

ViT-Tiny P8/P4 use token grids of $8\times8$/$16\times16$ and an embed dimension of 192. Their parameter counts are approximately 5.41M/5.42M, respectively, and the difference arises only from patch embedding and positional encoding. Downstream training uses AdamW with an initial learning rate of 5e$-$4, weight decay 0.05, a minimum learning rate of 1e$-$5, and 90 training epochs. All load-bearing ImageNet-100, ViT, and native-224px results in the main text use 3 seeds. Proxy and downstream seeds are paired, so the error bars cover end-to-end variation from selection through training. The ImageNet-1k low-resolution ladder and censored lower bounds use a single seed; a second seed only verifies the ordering of 32@32/32@64 at m=40,100.

\subsection{BN-Safe Budgets and Exact Subset Sizes}
\label{app:e3}

With batch 256, an extremely small final batch makes batch-normalization statistics unstable. We require that the remainder when the total selected sample count is divided by 256 be at least 16. The densified W2 budgets use exact subset sizes of 9000/11000, with remainders of 40/248. Every cell in the native-224px capacity sweep has a remainder of at least 20.

To satisfy the budget and BN constraints, we truncate only the preselected within-class order to the exact total sample count. The fixed R18 features, greedy coverage criterion, and comparison target remain unchanged.

\subsection{Measurement Status of $m_0$}
\label{app:e4}

For each budget m, we first compute $\mathrm{gap}=\mathrm{acc}(\text{easy-first})-\mathrm{acc}(\text{Herding})$ for each seed, then average under the seed protocol stated for that experiment and determine the sign. If two adjacent measured budgets exhibit a positive$\to$negative flip, we linearly interpolate between them to obtain $m_0$ and record the two endpoints as the measurement bracket. The measurement bracket describes budget discretization, not a statistical confidence interval.

If the gap remains positive throughout the measured range, we report only the right-censored bound that $m_0$ exceeds the maximum budget. If the gap is already negative at the minimum budget, we report only the left-censored bound that $m_0$ is below the minimum budget. Extrapolated values are not reported as results. For cells with a very shallow slope or local non-monotonicity, we prioritize the direction, measurement bracket, or one-sided lower bound and do not use spurious point-estimate precision from automatic interpolation. The operational criterion for a ``shallow crossing'' is that the $|\mathrm{gap}|$ values at the adjacent measured budgets on either side of the flip are of the same order as their cross-seed standard deviations. If the signs at the budgets adjacent to the flip are locally non-monotonic, we widen the measurement bracket outward by one measured budget before reporting it. Accordingly, the $(2\times,32)$ cell uses [80,110] rather than the mechanical [90,100] (Appendix~\ref{app:b2}).

\begin{table}[H]
\centering
\caption{\textbf{Evidence summary and measurement status.} ``Measured'' requires a positive-to-negative flip between adjacent budgets; ``censored'' reports only the boundary of the measured range; control experiments report only performance relations. a@b denotes a px of retained information on a b-px input grid (\S3.2), and R18/W2 denote ResNet-18 at $1\times/2\times$ width. Corroborating evidence such as TinyImageNet does not enter the load-bearing summary. The full-scale grid axis and capacity axis are listed separately, to avoid splicing different stem families onto one numerical axis.}
\label{tab:evidence}
\scriptsize
\setlength{\tabcolsep}{3pt}
\begin{tabularx}{\textwidth}{>{\raggedright\arraybackslash}p{0.10\textwidth}>{\raggedright\arraybackslash}p{0.17\textwidth}>{\raggedright\arraybackslash}p{0.07\textwidth}>{\raggedright\arraybackslash}X>{\raggedright\arraybackslash}X}
\toprule
Evidence & Setting & Seeds & $m_0$ / verdict & Status \\
\midrule
IN-100 input grid & 32@32/32@64/64@64, R18, $3\times3$-stem & 3 (paired) & 85 / 57 / 53 & measured \\
Frozen-list & 32@64 cell, R18 subsets $\to$ W2 & 3 & 57 $\to$ 85 & measured \\
Grid transplant ($2\times2$) & 32@32$\rightleftarrows$32@64 both directions, both arms' subsets frozen & 3 & learner main effect 18.3/18.4, list 9.3/9.4 & measured (control) \\
Capacity$\times$grid (LFrac) & $\{0.5\times,1\times,2\times\}\times\{32,64\}$ & 3 & grid 32: $\sim108$ (shallow)/85/$\sim95$ (shallow); grid 64: 64/57/85 & measured \\
Capacity$\times$grid (EL2N) & same as above & 3 & grid 32: $\ge100$/87/82; grid 64: 69/54/87 & 5 cells measured, 1 cell shallow-crossing lower bound \\
64@64 width sweep (LFrac) & $\{0.5\times,1\times,2\times\}$ at 64@64 & 3 (paired) & 57.1/53/59.1, spread about 6 & measured (shallow crossings at both ends, brackets overlap) \\
ViT four groups & 32@64, patch 8/4 $\times$ LFrac/EL2N & 3 & no crossing; $<25$ by convention & left-censored \\
ViT transplant control & R18 easy set $\to$ ViT & 3 & $\approx$Uniform ($+0.1$ to $+0.9$) & control: performance relations only \\
IN-1k input-grid ladder & 32@32/32@64/32@128, $3\times3$-stem & 1 & $>250$ / $>250$ / $>180$ & right-censored (a second seed rechecks m=40, 100) \\
IN-1k native width sweep & $0.5\times$--$2\times$, $7\times7$-stem, LFrac/EL2N & 3 & LFrac: 43.8/42.5/44.4; EL2N: 31.7/34.0/38.4 & measured (three widths share one measurement bracket within each probe) \\
Data side (cited) & sample pool shrunk $8.4\times$ / bandwidth compressed $49\times$ & -- & pool stability; modest bandwidth drift & prior work, not re-measured here \citep{liu2026dataside} \\
\bottomrule
\end{tabularx}
\end{table}

\setcounter{table}{0}\renewcommand{\thetable}{F\arabic{table}}\renewcommand{\theHtable}{F\arabic{table}}
\section{Run Inventory}
\label{app:inventory}

Each result file is one downstream training run (one selector, one learner, one budget, one seed). The main-text numbers of every experiment group, namely $m_0$ crossings, per-budget gaps, and accuracy ceilings, were re-aggregated from these files and checked against reference values with zero verification failures. Data-side results quoted from \citet{liu2026dataside} are cited, not re-verified here, and are outside the table's coverage.

\begin{table}[H]
\centering
\small
\begin{tabular}{llr}
\toprule
Experiment group & Datasets & Result files \\
\midrule
Input-grid ladder, native cells (32@32 / 32@64 / 64@64) & ImageNet-100 & 171 \\
Grid-ladder corroboration & TinyImageNet & 34 \\
Capacity$\times$grid map ($0.5\times$/$1\times$/$2\times$, incl.\ 64@64 wings) & ImageNet-100 & 378 \\
Cross-selector and cross-grid transplants & ImageNet-100 & 165 \\
Stride controls (layer-2, layer-4, and stem) & ImageNet-100 & 171 \\
Inductive bias (ViT-Tiny P8/P4) & ImageNet-100 & 221 \\
EL2N probe replication (six cells) & ImageNet-100 & 162 \\
Uniform reference arm (all ResNet cells) & ImageNet-100 & 189 \\
Full-scale low-resolution grid ladder & ImageNet-1k & 132 \\
Native-224px crossover baseline & ImageNet-1k & 78 \\
Native-224px width sweep (LFrac and EL2N probes) & ImageNet-1k & 138 \\
\midrule
Total & & 1839 \\
\bottomrule
\end{tabular}
\end{table}

\end{document}